\documentclass[10pt,twocolumn,letterpaper]{article}
\pdfoutput=1

\usepackage{cvpr}
\usepackage{indentfirst}

\definecolor{cvprblue}{rgb}{0.21,0.49,0.74}
\usepackage[pagebackref,breaklinks,colorlinks,allcolors=cvprblue]{hyperref}

\usepackage{graphicx}
\usepackage{rotating}
\usepackage{colortbl}  
\usepackage{booktabs}
\usepackage{multirow}
\usepackage{xcolor}
\usepackage{subcaption}
\usepackage{stfloats}

\title{$S^3$: Synonymous Semantic Space for Improving Zero-Shot Generalization of Vision-Language Models}

\author{Xiaojie Yin \quad Qilong Wang \quad Bing Cao \quad Qinghua Hu\\
Tianjin University\\
{\tt\small \{xjyin, qlwang, caobing, huqinghua\}}@tju.edu.cn}

\begin{document}
\maketitle
\begin{abstract}
Recently, many studies have been conducted to enhance the zero-shot generalization ability of vision-language models (e.g., CLIP) by addressing the semantic misalignment between image and text embeddings in downstream tasks. Although many efforts have been made, existing methods barely consider the fact that a class of images can be described by notably different textual concepts due to well-known lexical variation in natural language processing, which heavily affects the zero-shot generalization of CLIP. Therefore, this paper proposes a \textbf{S}ynonymous \textbf{S}emantic \textbf{S}pace ($S^3$) for each image class, rather than relying on a single textual concept, achieving more stable semantic alignment and improving the zero-shot generalization of CLIP. Specifically, our $S^3$ method first generates several synonymous concepts based on the label of each class by using large language models, and constructs a continuous yet compact synonymous semantic space based on the Vietoris-Rips complex of the generated synonymous concepts. Furthermore, we explore the effect of several point-to-space metrics on our $S^3$, while presenting a point-to-local-center metric to compute similarity between image embeddings and the synonymous semantic space of each class, accomplishing effective zero-shot predictions. Extensive experiments are conducted across 17 benchmarks, including fine-grained zero-shot classification, natural distribution zero-shot classification, and open-vocabulary segmentation, and the results show that our $S^3$ outperforms state-of-the-art methods.
\end{abstract}    
\section{Introduction}
\label{sec:intro}

With the aid of huge-scale training data of image-text pairs~\cite{radford2021learning,schuhmann2021laion,schuhmann2022laion}, pre-trained vision-language models (e.g., CLIP~\cite{radford2021learning}) have demonstrated promising zero-shot generalization ability. These vision-language models~\cite{radford2021learning,jia2021scaling,zhai2022lit,ge2022dall,ramesh2021zero} are good at understanding textual concepts involved in images, and therefore performing zero-shot classification by directly comparing embeddings of input images and those of textual labels in downstream tasks. However, there often exists a domain gap between pre-training data and that in domain-specific downstream tasks~\cite{mehrabi2021survey,menon2022task,agarwal2021evaluating}, especially annotation mode of textual concepts. This intuitively leads to semantic misalignment between image and text in the feature space defined by pre-trained vision-language models (VLMs), limiting the zero-shot generalization ability.

\begin{figure}[t]
    \centering
    \begin{subfigure}[b]{0.4\columnwidth}
        \centering
        \includegraphics[width=\textwidth]{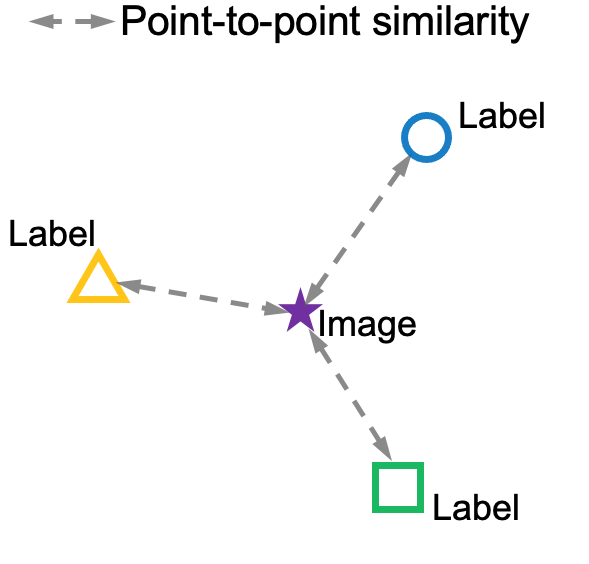}
        \caption{CLIP~\cite{radford2021learning}}
    \end{subfigure}
    \hspace{0.15cm}
    \begin{subfigure}[b]{0.4\columnwidth}
        \centering
        \includegraphics[width=\textwidth]{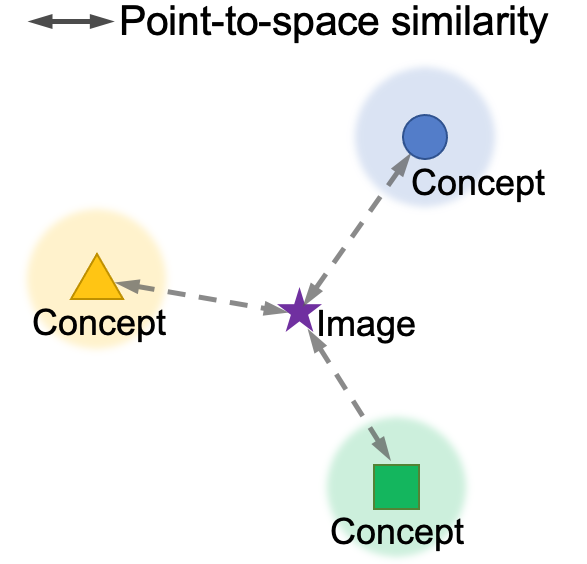}
        \caption{PE~\cite{pratt2023does,mirza2024meta}}
        \label{fig:pe}
    \end{subfigure}
    
    \vspace{0.1cm}
    
    \begin{subfigure}[b]{0.4\columnwidth}
        \centering
        \includegraphics[width=\textwidth]{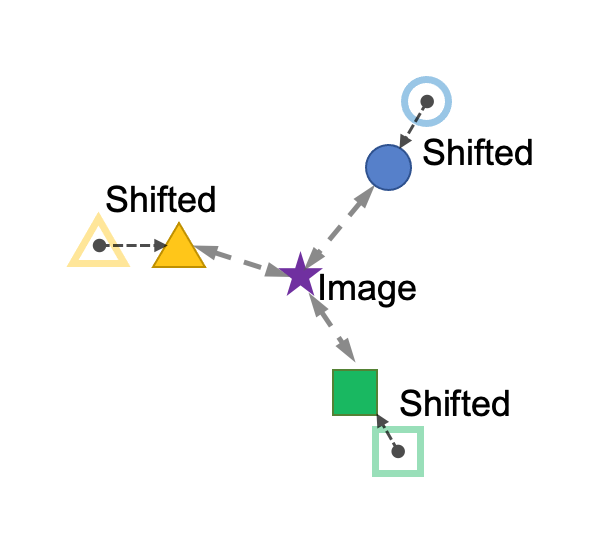}
        \caption{TTA~\cite{shu2022test,sui2024just}}
        \label{fig:tta}
    \end{subfigure}
    \hspace{0.15cm}
    \begin{subfigure}[b]{0.4\columnwidth}
        \centering
        \includegraphics[width=\textwidth]{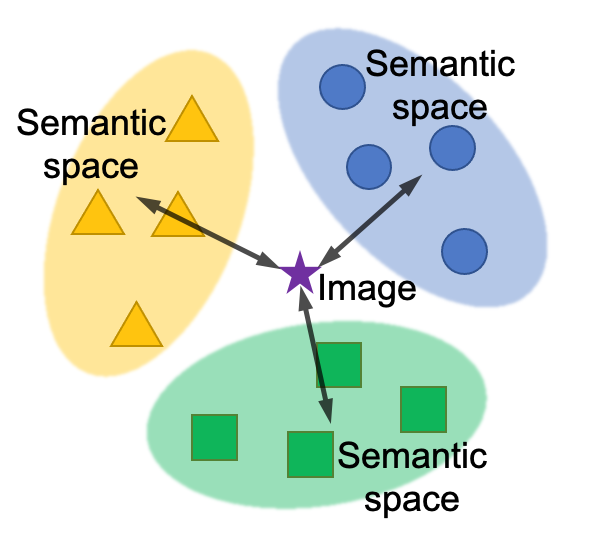}
        \caption{$S^3$ (Ours)}
    \end{subfigure}
    \caption{\textbf{Comparison of Methods.} (a) CLIP: Point-to-point similarity between image and label embeddings. (b) PE: Point-to-point similarity between image and single concept embeddings. (c) TTA: Point-to-point similarity between image and shifted text embeddings. (d) $S^3$ (Ours): Similarity between image and semantic spaces constructed from multiple synonymous concepts.}
\end{figure}

To address above issue, previous works can generally be divided into two categories: Prompt Engineering (PE) and Test-Time Adaptation (TTA). Specifically, as shown in Figure~\ref{fig:pe}, PE methods~\cite{menon2022visual,roth2023waffling,allingham2023simple,pratt2023does,mirza2024meta,parashar2024neglected} mainly focus on generating multiple detailed text descriptions for a single concept within each class by leveraging large language models (LLMs), e.g., GPT-4~\cite{achiam2023gpt} and Claude~\cite{anthropic2024claude}. Then, all descriptions are aggregated into a text embedding to represent the concept per class, which is used to compute similarity with the image embedding. As shown in Figure~\ref{fig:tta}, TTA methods~\cite{shu2022test,sui2024just,feng2023diverse,abdul2024align,ma2024swapprompt,zanella2024test} generally aim to dynamically adjust embeddings of text labels for each test sample during inference, and match the shifted embeddings of image-text pairs to alleviate issue of semantic misalignment.

\begin{figure*}[t]
    \centering
    \begin{subfigure}[t]{0.65\textwidth}
        \centering
        \includegraphics[width=\textwidth]{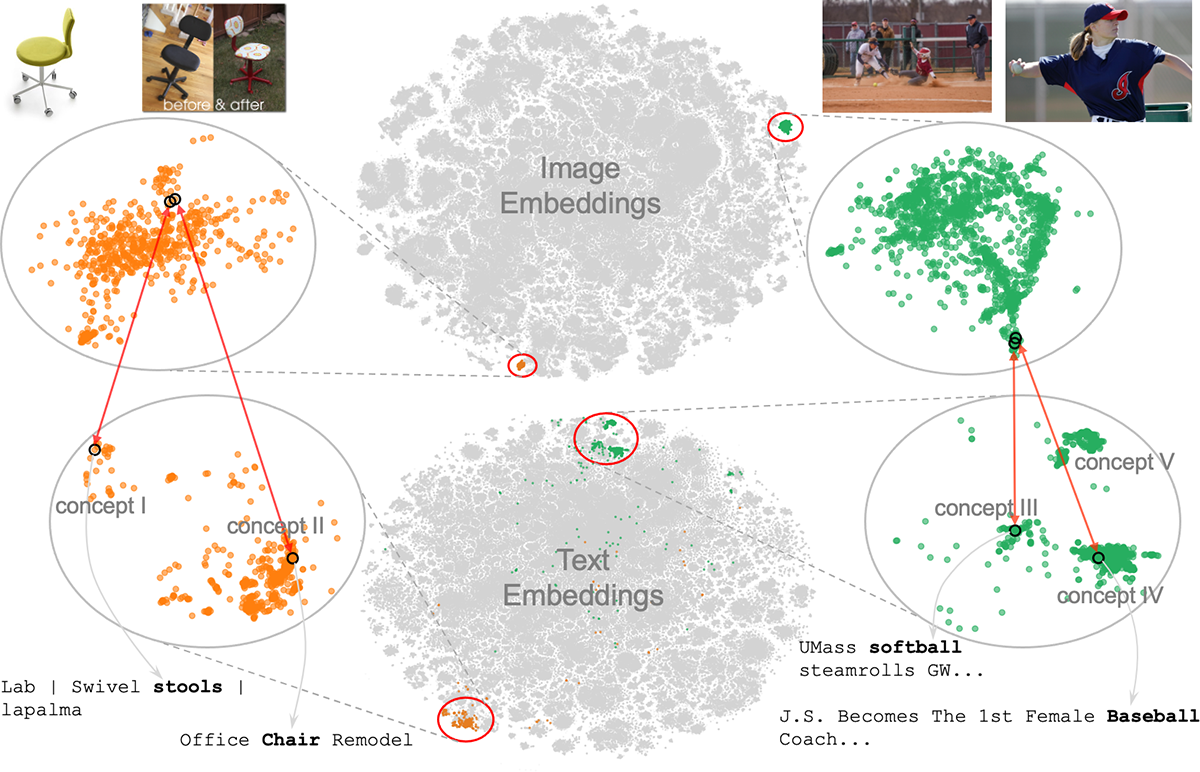}
        \caption{Lexical variation in LAION-400M dataset.}
        \label{fig:laion}
    \end{subfigure}
    \hspace{0.03\textwidth}
    \begin{subfigure}[b]{0.3\textwidth} 
        \begin{subfigure}[t]{\textwidth}
            \centering
            \includegraphics[width=\textwidth]{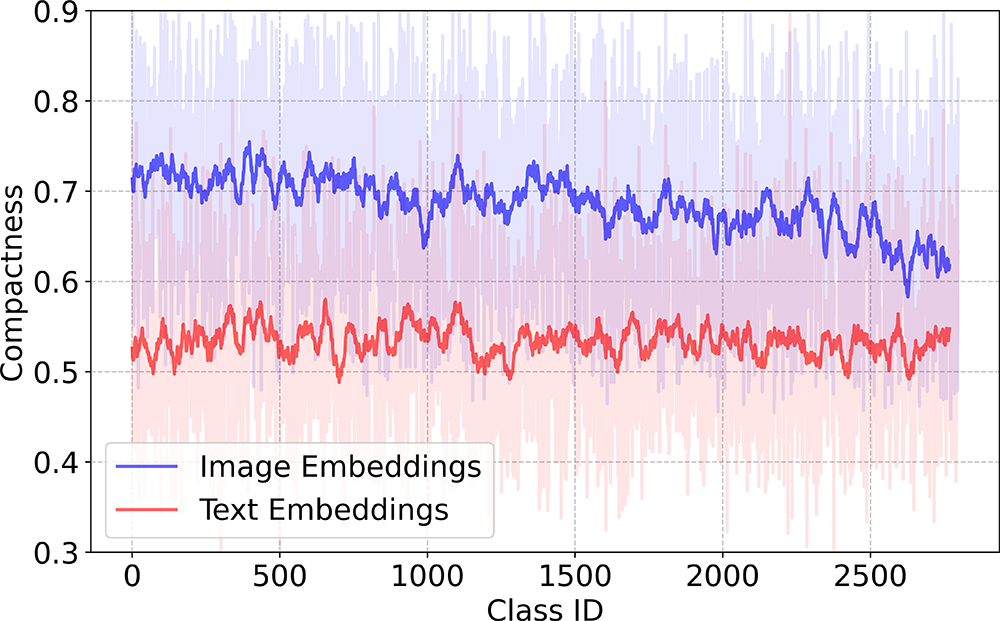}
            \caption{Compactness: image v.s. text.}
            \label{fig:trace}
        \end{subfigure}
        \vspace{0.1em}
        \vfill
        \begin{subfigure}[b]{\textwidth} 
            \centering
            \includegraphics[width=\textwidth]{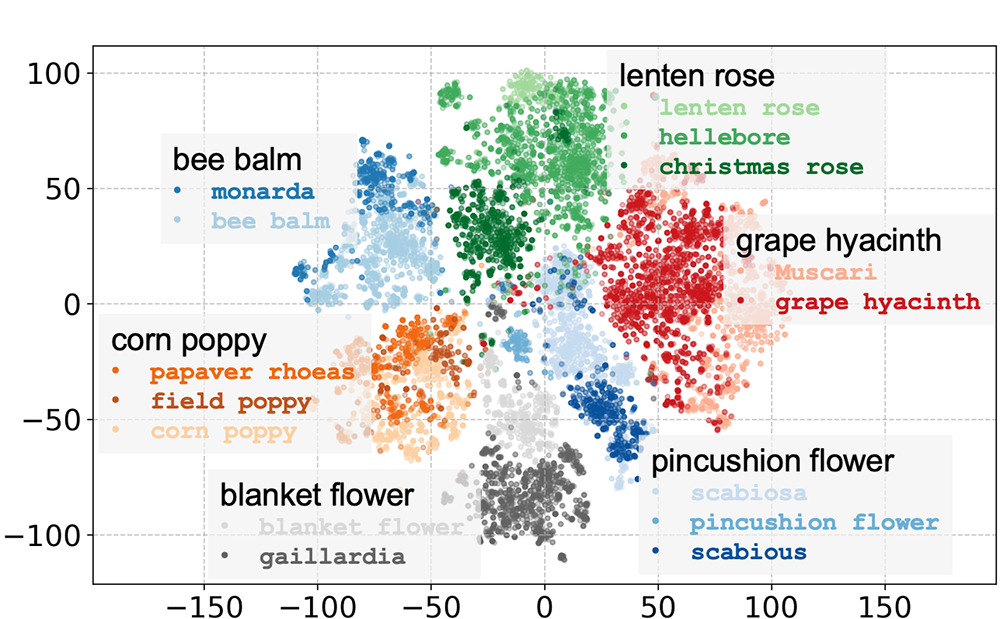}
            \caption{Synonymous texts form semantic spaces.}
            \label{fig:synonyms}
        \end{subfigure}
    \end{subfigure}
    \caption{\textbf{(a) Lexical variation in LAION-400M dataset:} Images of the same class with very similar visual embeddings correspond to significantly different text embeddings, which may even belong to different textual concepts. \textbf{(b) Compactness: image v.s. text:} Image embeddings (blue) are consistently more compact than text embeddings (red). The original data (light color) has been smoothed. \textbf{(c) Synonymous concepts form semantic spaces:} Different synonymous concepts for a class form continuous, non-overlapping spaces. }
    \label{fig:motivation}
\end{figure*}

Although significant progress has been made, previous works have primarily focused on aligning a class of images with a single textual concept. However, a well-known challenge in lexical variation within Natural Language Processing (NLP) is that a single concept can be expressed in multiple ways~\cite{parashar2024neglected}, and thus, a class of images represented by a single textual concept is potentially limited. To further analyze the effect of lexical variation on VLMs, we take huge-scale pre-training datasets of VLMs (i.e., LAION~\cite{schuhmann2021laion}) as an example. As shown in Figure~\ref{fig:laion}, we observe that images of the same class with very similar visual embeddings correspond to significantly different text embeddings, which may even belong to different textual concepts (e.g., `swivel stools' and `office chair remodel'). Furthermore, Figure~\ref{fig:trace} shows that the space of image embeddings is generally more compact than their text counterparts (refer to Sec. \ref{sec:observations} for more details). The observations above lead to the conclusion that a class of images is hardly comprehensively described by a single textual concept. Additionally, as illustrated in Figure \ref{fig:synonyms}, pre-trained VLMs naturally align similar textual descriptions~\cite{radford2021learning,abdul2024align}, and synonymous textual concepts merely form continuous, non-overlapping spaces in the embedding space.

Based on the above observations, this paper proposes a \textbf{S}ynonymous \textbf{S}emantic \textbf{S}pace ($S^3$) for describing each image class instead of a single textual concept, where a class of images is aligned with a space of textual concepts to better cope with lexical variation, further improving the zero-shot generalization ability of VLMs. To this end, our $S^3$ method first generates several different synonymous concepts and various detailed textual descriptors by providing existing large language models (e.g., GPT-4~\cite{achiam2023gpt}, Claude~\cite{anthropic2024claude}) with the label of each class, which are then combined to form a series of synonymous texts. Furthermore, we construct a continuous and compact synonymous semantic space by seeking the largest connected component in the topological properties of the semantic space. Specifically, we build a Vietoris-Rips complex~\cite{mischaikow2004computational,zhu2013persistent} for embeddings of the generated synonymous texts, which filters out noisy texts potentially resulting from hallucinations~\cite{zhang2023siren,huang2023survey} by large language models (LLMs) and forms a compact synonymous semantic space based on persistent homology. To accomplish zero-shot prediction, we explore several point-to-space metrics to calculate similarities between embeddings of test images and the synonymous semantic space of each class. In particular, we introduce a point-to-local-center metric that employs the center points of local regions nearest to image embeddings as the representative points of the semantic space, providing an efficient and stable similarity metric. The overview of our $S^3$ method is illustrated in Figure~\ref{fig:main}. To evaluate the effectiveness of our method, experiments are conducted on ten fine-grained zero-shot classification tasks (i.e., Flowers102~\cite{nilsback2008automated}, DTD~\cite{cimpoi2014describing}, Oxford Pets~\cite{parkhi2012cats}, Stanford Cars~\cite{krause20133d}, UCF101~\cite{soomro2012ucf101}, Caltech101~\cite{fei2004learning}, Food101~\cite{bossard2014food}, SUN397~\cite{xiao2010sun}, FGVC-Aircraft~\cite{maji2013fine}, and EuroSAT~\cite{helber2019eurosat}),  five natural distribution zero-shot datasets (e.g., ImageNet~\cite{deng2009imagenet}, ImageNet-A~\cite{hendrycks2021natural}, ImageNet-V2~\cite{recht2019imagenet}, ImageNet-R~\cite{hendrycks2021many}, and ImageNet-Sketch~\cite{wang2019learning}), and two open-vocabulary segmentation benchmarks (e.g., ADE20K~\cite{zhou2019semantic} and Pascal VOC~\cite{everingham2010pascal}). The contributions of this work are summarized as follows:


\begin{figure*}[t]
   \centering
   \includegraphics[width=0.9\linewidth]{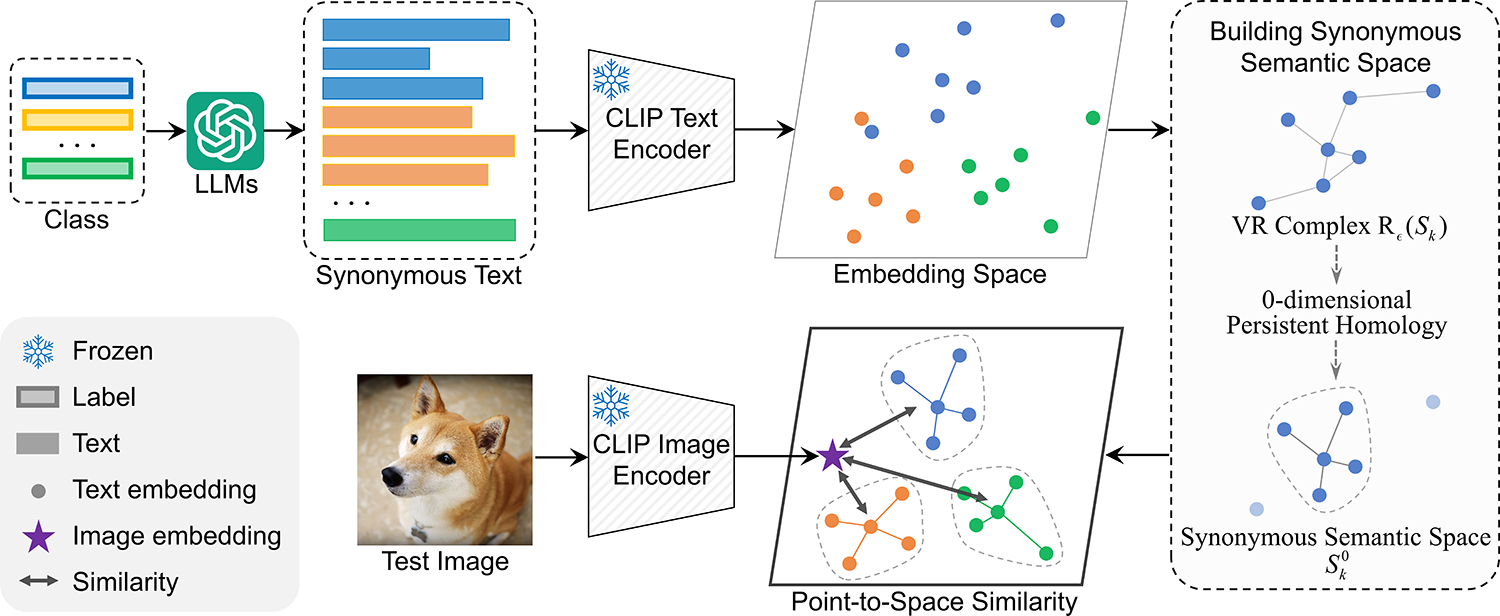}

   \caption{\textbf{Overall architecture of $S^3$.} Given label of each class, our $S^3$ method generates synonymous texts by prompting LLMs, which are used to construct a synonymous semantic space by seeking the largest connected component in topological properties of semantic space. For a test image, similarities between image embedding and synonymous semantic spaces are calculated for zero-shot prediction.}
   \label{fig:main}
\end{figure*}


\begin{itemize}
    \item To the best of our knowledge, this paper makes the first attempt to introduce the idea of a synonymous semantic space ($S^3$) to improve the zero-shot generalization of VLMs. Compared to a single text concept, our $S^3$ method better handles lexical variation in VLMs and achieves stable semantic alignment between the embeddings of image-text pairs.
    
    \item To this end, we construct a continuous yet compact synonymous semantic space based on LLMs by identifying the largest connected component in the Vietoris-Rips complex of synonymous text embeddings. Additionally, a point-to-local-center metric is introduced to provide an efficient and stable similarity metric between image embeddings and the synonymous semantic space for zero-shot prediction.

    \item  Extensive experiments are conducted across several benchmarks in fine-grained zero-shot classification, natural distribution zero-shot classification, and open-vocabulary segmentation. The results demonstrate that our $S^3$ method outperforms existing PE and TTA methods, achieving state-of-the-art performance.
\end{itemize}

\section{Related work}
\label{sec:related}

\noindent\textbf{Prompt Engineering (PE).} 
With the emergence of VLMs, prompt engineering has gained substantial attention in zero-shot learning. CLIP~\cite{radford2021learning} demonstrated that incorporating class names into human-engineered prompt templates significantly enhances classification accuracy. Building on this, ZPE~\cite{allingham2023simple} improves zero-shot performance by calculating the confidence by combining text with prompts. DCLIP~\cite{menon2022visual} extends this by using LLMs to generate textual descriptors of labels. WaffleCLIP~\cite{roth2023waffling} further enhances classification performance by incorporating random characters as descriptions in the prompt. CuPL~\cite{pratt2023does} and MPVR~\cite{mirza2024meta} directly leverage LLMs to generate prompts, achieving state-of-the-art performance. Additionally, REAL~\cite{parashar2024neglected} seeks to enhance effectiveness by replacing given labels with their most common synonyms identified through LLMs and open-source pre-trained datasets. In summary, PE generates multiple detailed text descriptions for a single concept within each class by leveraging LLMs, , which are then aggregated into a text embedding to represent the concept per class. However, this singular concept does not address the challenge of textual variation in CLIP. Our $S^3$ proposes a synonymous semantic space for describing each image class with multiple textual concepts, further improving the zero-shot generalization ability of VLMs. 

\noindent\textbf{Test-Time Adaptation (TTA).} 
TTA is a dynamic strategy employed during the testing phase to enhance a model's performance on specific tasks or data distributions. TPT~\cite{shu2022test} is the first to integrate TTA with zero-shot generation, adjusting prompts during testing. Building on TPT, DiffTPT~\cite{feng2023diverse} utilizes diffusion models to adjust image embeddings at test time. PromptAlign~\cite{abdul2024align} fine-tunes both text and image encodings using a proxy dataset during testing. SwapPrompt~\cite{ma2024swapprompt} and MTA~\cite{zanella2024test} focus on discovering more effective text-image matching patterns. Recently, TPS~\cite{sui2024just} achieved state-of-the-art performance by shifting text embeddings during testing. Shifting text or image embeddings alleviates issue of semantic misalignment but does not fully address the lexical variations in CLIP.

\section{Proposed method}
\label{sec:method}

In this section, we first discuss observations regarding image-text alignment in VLMs, which encourage us to propose a synonymous semantic space ($S^{3}$) method to improve zero-shot generalization of VLMs. As illustrated in Figure \ref{fig:main}, $S^{3}$ involves construction of synonymous semantic space and point-to-space similarity measure, whose details are given in Sec.~\ref{sec:construction} and Sec.~\ref{sec:similarity}, respectively. Finally, we briefly discuss how to integrate our $S^{3}$ into TTA method.


\begin{figure}[t]
   \centering
   \includegraphics[width=\columnwidth]{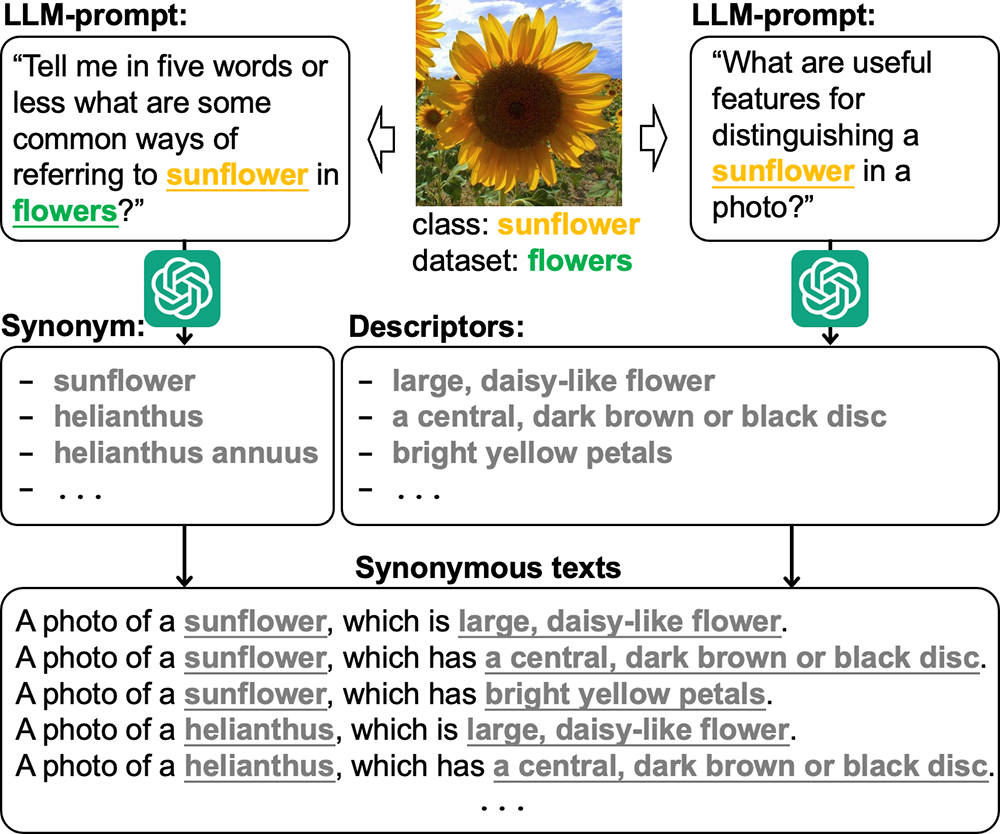}

   \caption{\textbf{Generating Synonymous Texts.} A class name (e.g., ``sunflower'') and its dataset name (e.g., ``flowers'') are given as inputs to the LLMs through two prompts. The first generates synonyms (e.g., ``sunflower'', ``helianthus''), and the second provides descriptors (e.g., ``large, daisy-like flower''). These are then combined into synonymous texts (e.g., ``A photo of a sunflower, which is a large, daisy-like flower''). }
   \label{fig:generation}
\end{figure}


\subsection{Observations on Image-Text Alignment in VLMs}
\label{sec:observations}

\noindent\textbf{Pre-trained VLMs Face Lexical Variation.} 
As a well-known challenge in NLP, lexical variation shows a single concept can be expressed in various ways~\cite{parashar2024neglected}. To further analyze the effect of lexical variation on pre-trained VLMs, we take huge-scale pre-training datasets of VLMs (i.e., LAION~\cite{schuhmann2021laion}) as an example. As shown in Figure~\ref{fig:laion}, we observe that images of the same class with very similar visual embeddings correspond to significantly different text embeddings, which may even belong to different textual concepts. For instance, two samples categorized as \texttt{"chair"} (top left) have a distance of only 0.02 in the image embedding space; however, they are assigned to two distinct concept clusters (i.e., \texttt{"chair"} and \texttt{"stools"}) in text embeddings, with a larger distance of 0.37. Similarly, for two samples categorized as \texttt{"baseball"} (top right), the distance in image embeddings is 0.02, but they are associated with different concepts (i.e., \texttt{"baseball"} and \texttt{"softball"}), resulting in a large distance of 0.30 between text embeddings. Furthermore, we analyze the compactness of image and text embeddings across 2,769 visual classes obtained through clustering. For the $i$-th cluster,  we compute the compactness of image embeddings by $1-\rm Tr(\boldsymbol\Sigma_i^{I})$, where $\rm Tr(\boldsymbol\Sigma_i^{I})$ denotes the trace of covariance matrix of all image embeddings in the $i$-th cluster. Intuitively, higher values of $1-\rm Tr(\boldsymbol\Sigma_i^{I})$ indicate more compactness. The same operation is also performed for text embeddings. As shown in Figure \ref{fig:trace}, the space of image embeddings (in blue) is generally more compact than their text counterparts (in red) with an average of 0.69 vs. 0.53. The above observations conclude that a class of images is hardly described by a single textual concept comprehensively.

\noindent\textbf{Synonymous Concepts in Downstream Tasks Form Semantic Spaces.} 
Previous works show that pre-trained VLMs can align semantically similar textual descriptions~\cite{radford2021learning, abdul2024align}, and here we investigate this phenomenon in downstream tasks. In this work, we take Flowers102~\cite{nilsback2008automated} as example and randomly select six categories. Then, we search the captions containing synonyms on labels of the selected six categories from the LAION-400M~\cite{schuhmann2021laion} dataset. The text embeddings of the corresponding captions are visualized in Figure \ref{fig:synonyms}, where we observe that the captions for each class (e.g., `lenten rose') consist of distinct regions, and each region corresponds to a synonymous concept (e.g., \texttt{"lenten rose"} in light green, \texttt{"hellebore"} in medium green, \texttt{"christmas rose"} in dark green). Particularly, these regions form a continuous yet non-overlapping semantic space for each class.


\subsection{Construction of Synonymous Semantic Space}
\label{sec:construction}

Above observations encourage us to construct a Synonymous Semantic Space ($S^3$) for describing each image class. However, construction of $S^3$ via label-to-caption retrieval in pre-training dataset raises up several challenges. Firstly, numerous class labels and their synonyms usually lack the corresponding captions in pre-training dataset, resulting in an incomplete semantic space. Secondly, substantial noise in pre-training dataset~\cite{yang2023alip} leads to the outliers in semantic space, bringing the side effect on zero-shot generalization in downstream tasks. To overcome above challenges, our generate diverse synonymous texts with prompting powerful LLMs, and construct a synonymous semantic space by seeking the largest connected component in topological properties of all generated synonymous texts.

\noindent\textbf{Generating Diverse Synonymous Texts.} 
To ensure the generated synonymous texts are as comprehensive as possible, we generate synonyms and descriptors of the labels separately, and then combine them. Given a class name and its dataset name, as shown in Figure~\ref{fig:generation}, we first prompt an off-the-shelf LLM (e.g. Claude~\cite{anthropic2024claude}) as: 
\texttt{"Tell me in five words or less what are some common ways of referring to \{class\} in \{dataset\}?"} 
It generates synonyms $\Phi_{\text{synonym}}(C_k)$ for the class $C_k$. Next, we query LLM to elicit descriptors of visual characteristics that effectively identify object categories in images: 
\texttt{"What are useful features for distinguishing a \{class\} in a photo?"} 
It aims to output distinguishing descriptors $\Phi_{\text{descriptor}}(C_k)$ for the class $C_k$. These synonyms and descriptors are combined through text template to produce synonymous texts $T_k$ as follows:
\begin{equation}
\begin{split}
& T_k = \mathrm{concatenate}\left(\phi_i , \phi_j \right), \\
& \forall \phi_i \in \Phi_{\text{synonym}}(C_k), \forall \phi_j \in \Phi_{\text{descriptor}}(C_k),
\end{split}
\end{equation}
where \(\mathrm{concatenate}\) refers to the operation of merging a synonym and descriptor according to the specified template \texttt{"A photo of a \{synonym\} which (is/has/etc) \{descriptor\}."}. 


\begin{figure}[t]
    \centering
    \begin{subfigure}[b]{0.24\columnwidth}
        \centering
        \includegraphics[width=0.99\textwidth]{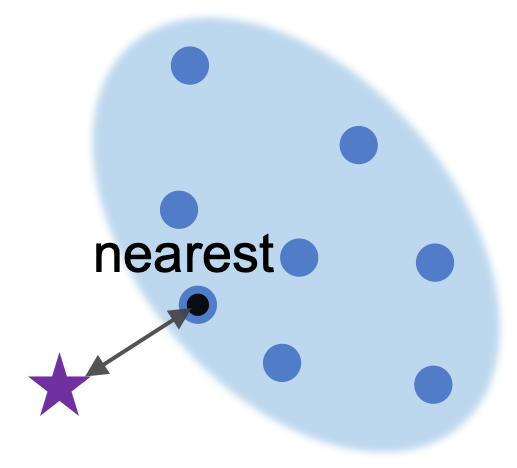}
        \caption{}
        \label{fig:set}
    \end{subfigure}
    \hfill
    \begin{subfigure}[b]{0.24\columnwidth}
        \centering
        \includegraphics[width=0.99\textwidth]{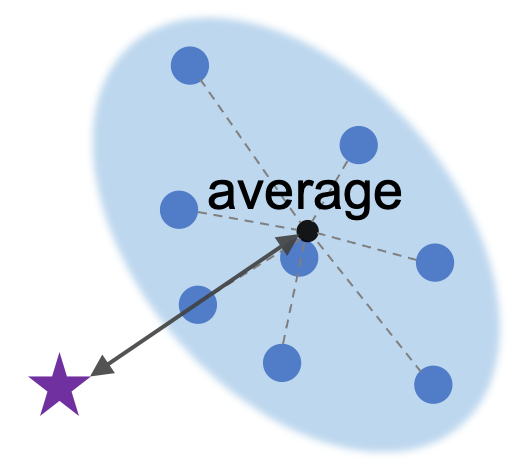}
        \caption{}
        \label{fig:center}
    \end{subfigure}
    \hfill
    \begin{subfigure}[b]{0.24\columnwidth}
        \centering
        \includegraphics[width=0.99\textwidth]{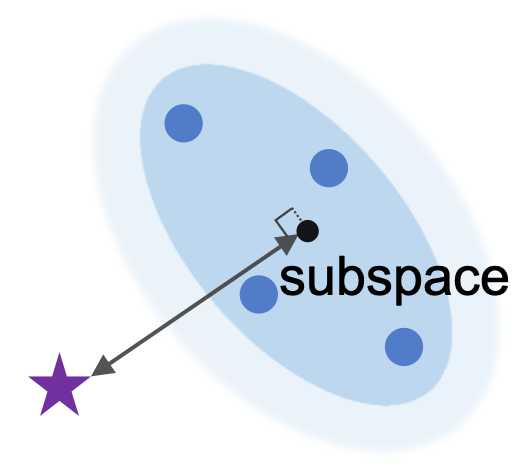}
        \caption{}
        \label{fig:subspace}
    \end{subfigure}
    \hfill
    \begin{subfigure}[b]{0.24\columnwidth}
        \centering
        \includegraphics[width=0.99\textwidth]{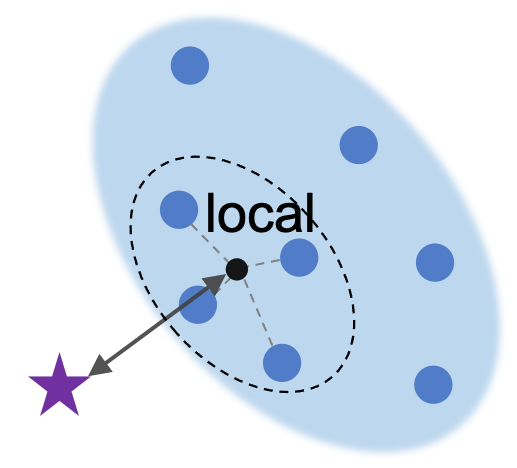}
        \caption{}
        \label{fig:local}
    \end{subfigure}
    \caption{\textbf{Point-to-Space Similarity Metric:} (a) Point-to-Set. (b) Point-to-Center. (c) Point-to-Subspace. (d) Point-to-Local-Center.}
    \label{fig:similiraty}
\end{figure}


\noindent\textbf{Compact Synonymous Semantic Spaces.} 
Next, we construct a compact synonymous semantic space based on the synonymous texts $T_k$. Specifically, given the generated synonymous texts \( T_k \), we employ the text encoder of CLIP 
\( \mathcal{F} \) to obtain the corresponding text embeddings, constructing the set \( S_k \) of synonymous text embeddings as follows:
\begin{equation}
S_k = \{ f_i \mid f_i = \mathcal{F}(t_i),\ t_i \in T_k \},
\end{equation}
which forms a synonymous semantic space for the class $C_k$. However, LLMs often produce hallucinations~\cite{zhang2023siren,huang2023survey}, and the generated content typically involves noise, which will affect the compactness of synonymous semantic space and bring the side effect on zero-shot generalization ability. To address this issue, we exploit the topological properties of the semantic space to identify the largest connected component, thereby filtering out noise data with weak semantic relevance and so guaranteeing the compactness of the synonymous semantic space. Specifically, we first construct a Vietoris-Rips complex~\cite{mischaikow2004computational,zhu2013persistent} for the set of \( S_k \):
\begin{equation}
\text{R}_\epsilon(S_k) = \left\{ \sigma \subseteq S_k \mid \langle f_i, f_j\rangle \geq \epsilon,\ \right. \left. \forall f_i, f_j \in \sigma \right\}.
\end{equation}
Here, \( \langle f_i, f_j \rangle \) represents the cosine similarity between embeddings \( f_i \) and \( f_j \), and \( \epsilon \) denotes the similarity threshold. As \( \epsilon \) increases, the connectivity within \( \text{R}_\epsilon(S_k) \) evolves, effectively capturing topological features across various scales. Subsequently, we apply 0-dimensional persistent homology to $\text{R}_\epsilon(S_k)$ to identify the largest connected component. Based on Topological Data Analysis (TDA)~\cite{wasserman2018topological}, we compute the birth $\epsilon_b(\gamma_i)$ and death $\epsilon_d(\gamma_i)$ times of each generator $\gamma_i$ in 0-dimensional persistent homology, with lifespan $\epsilon_d(\gamma_i) - \epsilon_b(\gamma_i)$ indicating persistence. As such, we identify the largest connected component $S_k^0$ at $\epsilon = \epsilon_{\text{max}}$, and $\epsilon_{\text{max}}$ corresponds to the generator's maximal lifespan: \begin{equation}
    S_k^0 = \bigcup_{\sigma_i \in \text{R}_{\epsilon=\epsilon_{max}}(S_k)} \sigma_i,
\end{equation}
where largest component $S_k^0$ provides a compact synonymous semantic space for the class $C_k$, exhibiting a better and more stable textual description than a single concept.

\subsection{Point-to-Space Similarity Metric}
\label{sec:similarity}

To accomplish zero-shot prediction, we require to measure the similarities between visual samples and synonymous semantic space of each class, instead of the original point-to-point similarities between visual samples and embedding vector of a single concept. Specifically, for a given test image \( I \), we compute its feature embedding \( g = \mathcal{G}(I) \) through the CLIP image encoder \( \mathcal{G} \), and then measure the similarity between $g$ and the synonymous semantic space \( S_k^0 \) of different classes. Finally, the class with the highest similarity score is identified as the predicted class.
\begin{equation}
y^* = \arg\max_{k} \mathrm{sim}(g, S_k^0).
\end{equation}

To measure the similarity between $g$ and synonymous semantic space \( S_k^0 \), we first introduce several point-to-space metrics as follows. As shown in Figure \ref{fig:set}, Point-to-Set metric~\cite{zhu2020progressive} computes similarity between the image embedding $g$ and the nearest neighbor in synonymous semantic space of $S_k^0$, which is degenerated into a point-to-point metric. Point-to-Center metric~\cite{menon2022visual} in Figure \ref{fig:center} computes similarity by comparing the image embedding $g$ with the centroid of $S_k^0$, which considers all information in each semantic space. Point-to-Subspace metric~\cite{turk1991eigenfaces}, as shown in Figure \ref{fig:subspace}, computes the similarity between $g$ and the mean of PCA basis of the embeddings in $S_k^0$, which projects image embeddings into the principal component subspace and computes the inner product with the mean of subspace. This metric quantifies the alignment between $g$ and the underlying structure of the semantic space.

\begin{table*}[t]
    \centering
    \footnotesize
    \setlength{\tabcolsep}{4pt}  
    \renewcommand{\arraystretch}{1.2}  
    \begin{tabular}{cl*{11}{p{0.9cm}}}
    \toprule
    & \textbf{Method} & \rotatebox{45}{\textbf{Flowers}} & \rotatebox{45}{\textbf{DTD}} & \rotatebox{45}{\textbf{Pets}} & \rotatebox{45}{\textbf{Cars}} & \rotatebox{45}{\textbf{UCF}} & \rotatebox{45}{\textbf{CalTech}} & \rotatebox{45}{\textbf{Food}} & \rotatebox{45}{\textbf{SUN}} & \rotatebox{45}{\textbf{Aircraft}} & \rotatebox{45}{\textbf{EuroSAT}} & \textbf{Avg.} \\
    \midrule
    \multirow{1}{*}{Baseline} & CLIP-ViT-B/16 & 67.28 & 44.44 & 87.98 & 65.24 & 65.08 & 92.98 & 83.80 & 62.55 & 23.70 & 41.41 & 63.45 \\
    \midrule
    \multirow{5}{*}{\parbox{1.7cm}{\centering Prompt \\ Engineering}}
    & DCLIP\dag~\cite{menon2022visual} & 70.52 & 49.82 & 87.30 & \textbf{66.70} & 70.34 & 93.96 & 84.50 & 67.47 & 24.81 & 44.37 & 65.98 \\
    & CuPL\dag~\cite{pratt2023does} & 73.57 & 49.17 & 91.25 & 66.10 & 70.31 & 93.96 & 84.44 & 67.66 & 27.84 & 50.70 & 67.50 \\
    & REAL\dag~\cite{parashar2024neglected} & 73.20 & 51.12 & 91.41 & 66.45 & 65.40 & 90.22 & 83.71 & 62.61 & 24.69 & 54.44 & 66.33 \\
    & MPVR~\cite{mirza2024meta} & 76.90 & \textbf{56.10} & 89.90 & 65.40 & \textbf{70.90} & \textbf{94.10} & \textbf{86.40} & \textbf{68.80} & 28.00 & 59.60 & 69.61 \\
    \rowcolor{gray!20}\cellcolor{white} & $S^3$ (Ours) & \textbf{81.36} & 53.96 & \textbf{91.58} & 66.45 & 70.39 & 93.59 & 84.02 & 67.77 & \textbf{29.73} & \textbf{61.51} & \textbf{70.04} \\
    \midrule
    \multirow{6}{*}{\parbox{1.7cm}{\centering Test-Time \\ Adaptation }}
    & TPT~\cite{shu2022test} & 68.98 & 47.75 & 87.79 & 66.87 & 68.04 & 94.16 & 84.67 & 65.50 & 24.78 & 42.44 & 65.10 \\
    & DiffTPT~\cite{feng2023diverse} & 70.10 & 47.00 & 88.22 & 67.01 & 68.22 & 92.49 & \textbf{87.23} & 65.74 & 25.60 & 43.13 & 65.47 \\
    & MTA~\cite{zanella2024test} & 68.06 & 45.90 & 88.24 & 68.47 & 66.69 & 94.21 & 85.00 & 66.67 & 25.20 & 45.36 & 65.58 \\
    & TPS~\cite{sui2024just} & 71.54 & 50.47 & 87.35 & \textbf{69.06} & 71.00 & \textbf{95.09} & 85.23 & \textbf{68.98} & 26.34 & 44.48 & 66.95 \\
    & \textcolor{gray}{OnZeta}~\cite{qian2024online}  & \textcolor{gray}{69.63} & \textcolor{gray}{48.58} & \textcolor{gray}{89.32} & \textcolor{gray}{69.03} & \textcolor{gray}{69.94} & \textcolor{gray}{93.89} & \textcolor{gray}{86.35} & \textcolor{gray}{69.01} & \textcolor{gray}{28.29} & \textcolor{gray}{56.74} & \textcolor{gray}{68.08} \\
    \rowcolor{gray!20}\cellcolor{white} & $TS^3$ (Ours) & \textbf{81.65} & \textbf{54.08} & \textbf{92.04} & 67.17 & \textbf{71.24} & 93.71 & 84.23 & 68.06 & \textbf{30.30} & \textbf{60.72} & \textbf{70.32} \\
    \bottomrule
    \end{tabular}
    \caption{Comparison with different state-of-the-art methods on ten zero-shot fine-grained classification benchmarks. Particularly, those methods with $\dag$ indicate that we reproduce them with same settings for fair comparison, and the \textbf{best accuracies} are highlighted in bold.}
    \label{tab:fg}
\end{table*}


\noindent\textbf{Point-to-Local-Center}. Different from the metrics above, we introduce a point-to-local-center metric by adaptively exploiting most relevant textual information to match image embeddings. Specifically, as shown in Figure~\ref{fig:local}, this metric first seeks a textual point nearest to \(g\) in \( S_k^0 \) indicated by $f^*$, and collects a local set consisting of $N$ points that are closest to $f^*$ in \( S_k^0 \), which is defined as \( N(f^*) \). Finally, we compute the similarity between the image embedding \(g\) and mean point of local set \( N(f^*) \) as
\begin{equation}\label{eqn-PLC}
\mathrm{sim}(g, S_k^0) = \langle g,  \frac{1}{|N(f^*)|} \sum_{f \in N(f^*)} f\rangle,
\end{equation}
where parameter $N$ is a hyperparameter that controls the neighborhood size. Compared to the point-to-set metric, point-to-local-center metric leverages more textual information, effectively improving the accuracy of text embeddings. In contrast with point-to-center and point-to-subspace metrics, point-to-local-center metric can eliminate interference from some textual embeddings those are not very relevant to \(g\). As such, our introduced point-to-local-center metric provides a more stable solution to align image embedding $g$ and synonymous semantic space \( S_k^0 \). 


\subsection{Test-Time $S^3$ Adaptation}
\label{sec:tta}


By considering the effect of downstream data on performance of zero-shot prediction, we introduce the idea of Test-Time Adaptation (TTA)~\cite{shu2022test,sui2024just} into our $S^3$, resulting in a $TS^3$ method. Specifically, given a test image $I$, we generate $M-1$ augmented images as suggested in~\cite{shu2022test,sui2024just}, and compute the embeddings $\{ g_i \}_{i=1}^{M}$ for both the original and augmented images by using CLIP image encoder. For each class, we apply a learnable vector $v_k$ to perform a uniform, channel-level shift in the synonymous semantic space $S_k^0$, yielding $S_k'$. Then, prediction probabilities for each $g_i$ are calculated by our point-to-local-center metric in Eqn.~(\ref{eqn-PLC}), while Top-$m$ distributions with the lowest entropy are used to compute the mean of prediction probabilities. By minimizing entropy of this marginal distribution, we can update $v_k$ via a single-step gradient descent. After adapting $v_k$ to $S_k^0$, we compute the similarity between the original image and $S_k'$ for zero-shot prediction. The proposed $TS^3$ method dynamically shifts the embeddings in synonymous semantic space for each test sample during inference, improving semantic alignment between image embeddings and synonymous semantic space and further enhancing zero-shot generalization.
\begin{table*}[t]
    \centering
    \footnotesize
    \setlength{\tabcolsep}{4pt}  
    \renewcommand{\arraystretch}{1.2}  
    \begin{tabular}{>{\centering\arraybackslash}p{2.2cm} *{1}{>{\arraybackslash}p{2.2cm}} *{6}{>{\centering\arraybackslash}p{1.7cm}}}
    \toprule
    & \textbf{Method} & \rotatebox{45}{\textbf{ImageNet}} & \rotatebox{45}{\textbf{-A}} & \rotatebox{45}{\textbf{-V2}} & \rotatebox{45}{\textbf{-R}} & \rotatebox{45}{\textbf{-Sketch}} & \textbf{Avg.} \\
    \midrule
    \multirow{1}{*}{Baseline} & CLIP-ViT-B/16 & 66.74 & 47.79 & 60.89 & 73.99 & 46.12 & 59.11 \\
    \midrule
    \multirow{4}{*}{\parbox{1.7cm}{\centering Prompt \\ Engineering}}
    & DCLIP\dag~\cite{menon2022visual} & 69.61 & 50.89 & 63.02 & 77.25 & 48.89 & 61.93 \\
    & CuPL\dag~\cite{pratt2023does} & 69.08 & \textbf{51.13} & 62.80 & 77.52 & 48.93 & 61.89 \\
    & REAL\dag~\cite{parashar2024neglected} & 68.50 & 50.04 & 61.97 & \textbf{77.69} & 48.19 & 61.28 \\
    \rowcolor{gray!20}\cellcolor{white} & $S^3$ (Ours) & \textbf{69.65} & 51.01 & \textbf{63.23} & 77.18 & \textbf{49.05} & \textbf{62.02} \\
    \midrule
    \multirow{5}{*}{\parbox{1.7cm}{\centering Test-Time \\ Adaptation }}
    & TPT~\cite{shu2022test} & 68.98 & 54.77 & 63.45 & 77.06 & 47.94 & 62.44 \\
    & DiffTPT~\cite{feng2023diverse} & 70.30 & 55.68 & \textbf{65.10} & 75.00 & 46.80 & 62.58 \\
    & MTA~\cite{zanella2024test} & 70.08 & 58.06 & 64.24 & 78.33 & 49.61 & 64.06 \\
    & TPS~\cite{sui2024just} & 71.45 & 60.61 & 64.91 & \textbf{80.20} & 50.88 & 65.61 \\
    \rowcolor{gray!20}\cellcolor{white} & $TS^3$ (Ours) & \textbf{71.57} & \textbf{61.11} & 65.04 & 80.06 & \textbf{50.96} & \textbf{65.75} \\
    \bottomrule
    \end{tabular}
    \caption{Comparison with state-of-the-arts on zero-shot natural distribution classification benchmarks.}
    \label{tab:nd}
\end{table*}

\begin{table}[t]
    \centering
    \footnotesize
    \setlength{\tabcolsep}{4pt} 
    \renewcommand{\arraystretch}{1.2}  
    \begin{tabular}{l*{2}{>{\centering\arraybackslash}p{0.9cm}}*{2}{>{\centering\arraybackslash}p{0.9cm}}*{1}{>{\centering\arraybackslash}p{0.9cm}}}
    \toprule
    \multirow{2}{*}{\textbf{Method}} & \multicolumn{2}{c}{\textbf{ADE20K}} & \multicolumn{2}{c}{\textbf{Pascal VOC}} & \multirow{2}{*}{\textbf{Avg.}} \\
    \cmidrule(lr){2-3} \cmidrule(lr){4-5} & \textbf{w/o BG} & \textbf{w/ BG} & \textbf{w/o BG} & \textbf{w/ BG} \\
    \midrule
    MaskCLIP+~\cite{zhou2022extract} & 9.12 & 8.53 & 47.59 & 26.17 & 22.85 \\
    + CuPL~\cite{pratt2023does} & 9.93 & 9.32 & 49.75 & 28.03 & 24.26 \\
    + REAL~\cite{parashar2024neglected} & 9.48 & 8.85 & 49.60 & 27.28 & 23.80 \\
    \rowcolor{gray!20} + $S^3$ (Ours) & \textbf{10.39} & \textbf{9.67} & \textbf{50.98} & \textbf{34.14} & \textbf{26.30} \\
    \midrule
    LSeg+~\cite{li2022language} & 31.35 & 28.09 & 74.92 & 50.50 & 46.22 \\
    + CuPL~\cite{pratt2023does} & 32.38 & 29.10 & 75.93 & 57.56 & 48.74 \\
    + REAL~\cite{parashar2024neglected} & 33.14 & 29.65 & 75.17 & 56.10 & 48.52 \\
    \rowcolor{gray!20} + $S^3$ (Ours) & \textbf{34.00} & \textbf{30.54} & \textbf{76.05} & \textbf{63.36} & \textbf{50.99} \\
    \bottomrule
    \end{tabular}
    \caption{Comparison of different methods on open-vocabulary segmentation task, where results of mIoU on ADE20K and Pascal VOC with and without background (BG) are reported.}
    \label{tab:ov}
\end{table}

\begin{table}[ht]
    \centering
    \footnotesize
    \setlength{\tabcolsep}{4pt}  
    \renewcommand{\arraystretch}{1.2}  
    \begin{tabular}{>{\centering\arraybackslash}p{2cm}*{1}{>{\arraybackslash}p{2.8cm}}*{1}{>{\centering\arraybackslash}p{2.4cm}}}
    \toprule
    & \textbf{Method} & \textbf{Avg.} \\
    \midrule
    \multirow{2}{*}{\parbox{1.7cm}{\centering Selection of \\ LLMs}}
    & GPT-4 & 68.88 \\
    & \cellcolor{gray!20}Claude & \cellcolor{gray!20}\textbf{70.04} \\
    \midrule
    \multirow{3}{*}{\parbox{1.7cm}{\centering Effect of \\ Homology }}
    & CLIP-ViT-B/16 & 63.45 \\
    & $S^3$ (w/o homology) & 68.50 \\
    & \cellcolor{gray!20} $S^3$ (w/ homology) & \cellcolor{gray!20}\textbf{70.04} \\
    \bottomrule
    \end{tabular}
    \caption{Ablation studies on selection of LLMs and effect of Homology.}
    \label{tab:ablation}
\end{table}


\section{Experiments}
\label{sec:experimental}

\subsection{Experimental Settings}
\noindent\textbf{Implementation Details.}
In this work, we adopt CLIP ViT-B/16 as the basic architecture to implement all methods for comparison. To generate the diverse synonymous texts, we build the script based on the code repository provided in \cite{menon2022visual}, while employing the public web API of Claude-3.5-Sonnet~\cite{anthropic2024claude} and GPT-4~\cite{achiam2023gpt} to generate synonyms and produce detailed descriptors, respectively. The hyperparameters of similarity threshold ($\epsilon_{\text{max}}$) and neighborhood size ($N$) are discussed in the supplementary materials. All experiments are conducted using PyTorch on a single NVIDIA RTX 3090 GPU. Source code will be publicly available. 

\noindent\textbf{Competing Methods.}
To evaluate our $S^3$ method, we compare with four PE methods (i.e., DCLIP~\cite{menon2022visual}, CuPL~\cite{pratt2023does}, REAL~\cite{parashar2024neglected}, and MPVR~\cite{mirza2024meta}), as well as five TTA methods (i.e., TPT~\cite{shu2022test}, DiffTPT~\cite{feng2023diverse}, MTA~\cite{zanella2024test}, TPS~\cite{sui2024just}, and OnZeta~\cite{qian2024online}). The baseline employs  CLIP of ViT-B/16 with standard prompt templates.  Particularly, we reproduce DCLIP, CuPL, and REAL by using CLIP of ViT-B/16 for fair comparison. For open-vocabulary segmentation, we employ CuPL, REAL and our $S^3$ as a replacement for the text embeddings in two widely used methods, including MaskCLIP+~\cite{zhou2022extract} and LSeg+~\cite{li2022language}.

\subsection{Results on Fine-Grained Datasets}

\noindent\textbf{Datasets.} We report top-1 accuracy on 10 fine-grained datasets including Flowers102~\cite{nilsback2008automated}, DTD~\cite{cimpoi2014describing}, Oxford Pets~\cite{parkhi2012cats}, Stanford Cars~\cite{krause20133d}, UCF101~\cite{soomro2012ucf101}, Caltech101~\cite{fei2004learning}, Food101~\cite{bossard2014food}, SUN397~\cite{xiao2010sun}, FGVC-Aircraft~\cite{maji2013fine}, and EuroSAT~\cite{helber2019eurosat}.

\noindent\textbf{Comparison with PE Methods.} As shown in Table \ref{tab:fg}, our method achieves the highest average accuracy of 70.04\%. Compared to DCLIP, which generates descriptors using LLMs, and REAL, which generates synonyms using LLMs, our method improves performance by $\sim$4\% and $\sim$3.7\% on average, respectively. When compared to CuPL and MPVR, which generate prompt texts using LLMs, our method shows average improvements of $\sim$2.5\% and $\sim$0.4\%, respectively. It is notable that  our method is much more cost-effective than MVPR, requiring only 6\% of MVPR's cost (see supplementary materials for details). These results clearly demonstrate the superiority of our synonymous semantic space over single generated semantic concept in semantic alignment between the embeddings of image-text pairs under the zero-shot setting.

\noindent\textbf{Comparison with TTA Methods.} 
As shown in Table \ref{tab:fg}, our $TS^3$ method achieves the highest average accuracy of 70.32\%. Compared to TPS, which shifts text embeddings in the DCLIP method, our method improves accuracy by $\sim$3.3\%. Additionally, our method outperforms the online learning-based OnZeta by $\sim$2.2\%. Despite TTA alleviates semantic misalignment by shifting embeddings, idea of synonymous semantic space can further bring clear improvement,  verifying the effectiveness of our $S^3$ again.


\subsection{Results on Natural Distribution Datasets}

\noindent\textbf{Datasets.} We report top-1 accuracy on 5 natural distribution datasets including ImageNet~\cite{deng2009imagenet} and its out-of-distribution variants ImageNet-A~\cite{hendrycks2021natural}, ImageNet-V2~\cite{recht2019imagenet}, ImageNet-R~\cite{hendrycks2021many}, and ImageNet-Sketch~\cite{wang2019learning}.

\noindent\textbf{Comparison with PE Methods.} 
As shown in Table \ref{tab:nd}, our method achieves the highest average accuracy of 62.02\%, surpassing most state-of-the-art baseline methods across various datasets, particularly on ImageNet. Compared to DCLIP and REAL, our method improves on all datasets except ImageNet-R, with an average improvement of $\sim$0.1\%. When compared to CuPL, our method shows an overall improvement of 0.13\%.

\noindent\textbf{Comparison with TTA Methods.} 
As shown in Table \ref{tab:nd}, our $TS^3$ method further improves performance, achieving an average accuracy of 65.75\%. Compared to the second-best TPS, our method shows an improvement of $\sim$0.1\%. The performance boost from TTA is particularly evident on difficult out-of-distribution datasets like ImageNet-A, where our method achieves 61.11\%, and ImageNet-S, where it reaches 50.96\%, showing strong robustness under various distribution shifts. They show good generalization of our $S^3$.


\begin{table*}[t]
    \centering
    \footnotesize
    \setlength{\tabcolsep}{4pt}  
    \renewcommand{\arraystretch}{1.2}  
    \begin{tabular}{p{2.5cm}*{11}{>{\centering\arraybackslash}p{1cm}}}
    \toprule
    \textbf{Metric} & \rotatebox{45}{\textbf{Flowers}} & \rotatebox{45}{\textbf{DTD}} & \rotatebox{45}{\textbf{Pets}} & \rotatebox{45}{\textbf{Cars}} & \rotatebox{45}{\textbf{UCF}} & \rotatebox{45}{\textbf{CalTech}} & \rotatebox{45}{\textbf{Food}} & \rotatebox{45}{\textbf{SUN}} & \rotatebox{45}{\textbf{Aircraft}} & \rotatebox{45}{\textbf{EuroSAT}} & \textbf{Avg.} \\
    \midrule
    Point-to-Set     & \textbf{81.36} & 50.77 & 89.26 & 65.51 & 68.86 & 92.49 & 83.75 & 63.69 & 27.99 & 57.91 & 68.16 \\
    Point-to-Center   & 75.36 & 53.37 & 91.36 & 66.43 & 70.37 & 93.47 & \textbf{84.02} & \textbf{67.77} & 29.67 & 61.49 & 69.32 \\
    Point-to-Subspace       & 74.54 & 53.37 & 91.52 & 66.43 & \textbf{70.61} & \textbf{93.59} & \textbf{84.02} & \textbf{67.77} & 29.67 & \textbf{61.64} & 69.31 \\
    \rowcolor{gray!20} Point-to-Local-Center  & \textbf{81.36} & \textbf{53.96} & \textbf{91.58} & \textbf{66.45} & 70.39 & \textbf{93.59} & \textbf{84.02} & \textbf{67.77} & \textbf{29.73} & 61.51 & \textbf{70.04} \\
    \bottomrule
    \end{tabular}
    \caption{Results of different point-to-space similarity metrics on zero-shot fine-grained classification benchmarks.}
    \label{tab:sim}
\end{table*}


\subsection{Results on Open-Vocabulary Segmentation}

\noindent\textbf{Datasets.} 
Open-vocabulary segmentation aims to understand an image with arbitrary categories described by texts. We conduct experiments on the challenging ADE20K~\cite{zhou2019semantic} and Pascal VOC~\cite{everingham2010pascal} datasets. ADE20K is a densely annotated dataset for scene understanding, comprising 150 categories and a background (BG). Pascal VOC is a classic dataset with 20 categories and a background (BG).

\noindent\textbf{Results and Analysis.} 
As shown in Table~\ref{tab:ov}, based on MaskCLIP+ and LSeg+ methods, our $S^3$ respectively achieves average mIoUs of 26.30\% and 50.99\%, significantly improving open-vocabulary segmentation performance. Notably, when considering background, our method provides gains of $\sim$1.1\% and $\sim$2.4\% on the ADE20K dataset, and $\sim$8\% and $\sim$12.9\% on the Pascal VOC dataset. Furthermore, we also compare with two PE methods, including CuPL and REAL. Compared to CuPL, our method achieves gains of $\sim$2\% and $\sim$2.2\% with MaskCLIP+ and LSeg+, respectively. Compared to REAL, our method achieves gains of $\sim$2.5\% and $\sim$2.4\%. These results show our $S^3$ can be well generalized to various zero-shot tasks.


\subsection{Ablation Study}

\noindent\textbf{Selection of LLMs.} 
We compare the performance of two LLMs for text generation, i.e., GPT-4 and Claude for synonymous text generation. As shown in Table \ref{tab:ablation} (top half). Claude consistently outperformed GPT-4 by an average of $\sim$1.2\% on all datasets. Consequently, we selected Claude for synonym generation.

\noindent\textbf{Effect of Homology.} 
To assess the influence of persistent homology in constructing synonymous semantic spaces, we conduct an ablation study \textit{w/} and \textit{w/o} homology. As highlighted in Table \ref{tab:ablation} (bottle half), integration of homology improves performance across all datasets, and achieves an average improvement of $\sim$1.5\%. This indicates that homology contributes to construct more compact semantic spaces, thereby enhancing zero-shot performance.

\noindent\textbf{Comparison of Point-to-Space Similarity Metrics.} 
We evaluated four point-to-space similarity metrics in Sec \ref{sec:similarity}: \textit{Point-to-Set}, \textit{Point-to-Center}, \textit{Point-to-Subspace}, and \textit{Point-to-Local-Center}. As shown in Table \ref{tab:sim}, the \textit{Point-to-Local-Center} similarity consistently outperforms the others across multiple datasets, achieving the highest average accuracy of 70.04\%. In comparison, the \textit{Point-to-Subspace} and \textit{Point-to-Center} metrics exhibit slightly weaker performance, with average accuracy of 69.31\% and 69.32\%, respectively. The \textit{Point-to-Set} similarity performs only showed exceptional performance on the Flowers dataset. These results clearly show \textit{Point-to-Local-Center} is a more stable and effective point-to-space similarity metric.

\section{Conclusions}


In this work, we propose the Synonymous Semantic Space ($S^3$) to address lexical variation in vision-language models, improving zero-shot generalization by representing each image class with a space of synonymous textual concepts. Our method outperforms existing approaches across multiple benchmarks, including fine-grained zero-shot classification, natural distribution zero-shot classification, and open-vocabulary segmentation. Future work will focus on further optimizing the $S^3$ method and exploring its applications in other tasks like cross-modal retrieval.
{
    \small
    \bibliographystyle{ieeenat_fullname}
    \bibliography{main}
}

\appendix 
\renewcommand{\thesection}{\Alph{section}}
\renewcommand{\thetable}{S\arabic{table}}
\renewcommand{\thefigure}{S\arabic{figure}}
\clearpage
\setcounter{page}{1}
\setcounter{table}{0}
\setcounter{figure}{0}
\maketitlesupplementary

\section{Analysis of Cost-Efficient}

As shown in Table \ref{tab:cost}, we compare the token costs incurred when using LLMs as generators of different methods. Compared to the second-best performing MPVR, our method achieves a 0.43\% increase in accuracy at only 6\% of MPVR's cost. This cost efficiency is due to our method querying LLMs for synonyms and descriptive phrases, which are much shorter than the rich visual prompts MPVR requires. For every 1K categories, MPVR generates 1000K tokens (costing \$10) using ChatGPT, while our method only requires 10K tokens (\$0.1) for synonyms and 50K tokens (\$0.5) for descriptors, totaling just \$0.6. Compared to CuPL, our method achieves a 2.54\% higher accuracy at only 12\% of its cost. Furthermore, compared to DCLIP and REAL, our method achieves significantly higher accuracy under similar costs, improving by 4.06\% and 3.71\%, respectively. These results highlight the remarkable cost-effectiveness of our method, achieving superior accuracy while maintaining significantly lower cost.

\begin{table}[ht]
    \centering
    \footnotesize
    \setlength{\tabcolsep}{4pt}  
    \renewcommand{\arraystretch}{1.2}  
    \begin{tabular}{p{1.5cm}*{5}{>{\centering\arraybackslash}p{1cm}}}
        \toprule
        \multirow{3}{*}{Model} & \multirow{3}{*}{Accuracy} & \multicolumn{3}{c}{LLMs Generator} & \multirow{3}{*}{Token} \\
        \cmidrule(lr){3-5}
        & \multirow{2}{*}{(\%)} & Words & Phrases & Sentences & \multirow{2}{*}{Cost} \\
        & & (10K) & (50K) & (500K) & \\
        \midrule
        DCLIP~\cite{menon2022visual} & 65.98 & - & 1 & - & 50K \\
        CuPL~\cite{pratt2023does}  & 67.50 & - & - & 1 & 500K \\
        REAL~\cite{parashar2024neglected} & 66.33 & 1 & - & - & 10K \\
        MPVR~\cite{mirza2024meta} & 69.61 & - & - & 2 & 1000K \\
        \rowcolor{gray!20} Ours & 70.04 & 1 & 1 & - & 60K \\
        \bottomrule
    \end{tabular}
    \caption{Token cost analysis across different PE methods. For every 1K categories, LLMs generate approximately 10K tokens for words, 50K tokens for phrases, and 500K tokens for sentences (refer to \cite{parashar2024neglected}).}
    \label{tab:cost}
\end{table}


\begin{table*}[t]
    \centering
    \footnotesize
    \setlength{\tabcolsep}{4pt}  
    \renewcommand{\arraystretch}{1.2}  
    \begin{tabular}{p{1.9cm}*{11}{>{\centering\arraybackslash}p{1.05cm}}}
    \toprule
    \textbf{Model} & \rotatebox{45}{\textbf{Flowers}} & \rotatebox{45}{\textbf{DTD}} & \rotatebox{45}{\textbf{Pets}} & \rotatebox{45}{\textbf{Cars}} & \rotatebox{45}{\textbf{UCF}} & \rotatebox{45}{\textbf{CalTech}} & \rotatebox{45}{\textbf{Food}} & \rotatebox{45}{\textbf{SUN}} & \rotatebox{45}{\textbf{Aircraft}} & \rotatebox{45}{\textbf{EuroSAT}} & \textbf{Avg.} \\
    \midrule
    GPT-4 & 76.70 & 53.90 & 91.41 & 64.98 & 69.84 & 92.86 & 79.80 & 67.76 & \textbf{30.03} & \textbf{61.53} & 68.88 \\
    \rowcolor{gray!20} Claude & \textbf{81.36} & \textbf{53.96} & \textbf{91.58} & \textbf{66.45} & \textbf{70.39} & \textbf{93.59} & \textbf{84.02} & \textbf{67.77} & 29.73 & 61.51 & \textbf{70.04} \\
    \bottomrule
    \end{tabular}
    \caption{Ablation study on selection of LLMs for synonymous texts generation across zero-shot fine-grained classification benchmarks.}
    \label{tab:llm}
\end{table*}

\begin{table*}[ht]
    \centering
    \footnotesize
    \setlength{\tabcolsep}{4pt}  
    \renewcommand{\arraystretch}{1.2}  
    \begin{tabular}{p{2.5cm}*{11}{>{\centering\arraybackslash}p{1cm}}}
    \toprule
    \textbf{Method} & \rotatebox{45}{\textbf{Flowers}} & \rotatebox{45}{\textbf{DTD}} & \rotatebox{45}{\textbf{Pets}} & \rotatebox{45}{\textbf{Cars}} & \rotatebox{45}{\textbf{UCF}} & \rotatebox{45}{\textbf{CalTech}} & \rotatebox{45}{\textbf{Food}} & \rotatebox{45}{\textbf{SUN}} & \rotatebox{45}{\textbf{Aircraft}} & \rotatebox{45}{\textbf{EuroSAT}} & \textbf{Avg.} \\
    \midrule
    CLIP-ViT-B/16 & 67.28 & 44.44 & 87.98 & 65.24 & 65.08 & 92.98 & 83.80 & 62.55 & 23.70 & 41.41 & 63.45 \\
    Ours (w/o homology)   & 79.13 & 52.84 & 89.40 & 66.37 & 70.29 & 93.27 & 84.00 & 67.68 & 29.25 & 52.79 & 68.50 \\
    \rowcolor{gray!20} Ours (w/ homology)  & \textbf{81.36} & \textbf{53.96} & \textbf{91.58} & \textbf{66.45} & \textbf{70.39} & \textbf{93.59} & \textbf{84.02} & \textbf{67.77} & \textbf{29.73} & \textbf{61.51} & \textbf{70.04} \\
    \bottomrule
    \end{tabular}
    \caption{Ablation study on effect of Homology across zero-shot fine-grained classification benchmarks.}
    \label{tab:homology}
\end{table*}

\begin{table*}[!ht]
    \centering
    \footnotesize
    \setlength{\tabcolsep}{4pt}  
    \renewcommand{\arraystretch}{1.2}  
    
    \begin{tabular}{>{\centering\arraybackslash}p{1.1cm} *{1}{>{\arraybackslash}p{1.1cm}} *{11}{>{\centering\arraybackslash}p{1cm}}}
    \toprule
    \textbf{Arch} & \textbf{Method} & \rotatebox{45}{\textbf{Flowers}} & \rotatebox{45}{\textbf{DTD}} & \rotatebox{45}{\textbf{Pets}} & \rotatebox{45}{\textbf{Cars}} & \rotatebox{45}{\textbf{UCF}} & \rotatebox{45}{\textbf{CalTech}} & \rotatebox{45}{\textbf{Food}} & \rotatebox{45}{\textbf{SUN}} & \rotatebox{45}{\textbf{Aircraft}} & \rotatebox{45}{\textbf{EuroSAT}} & \textbf{Avg.} \\
    \midrule
    \multirow{2}{*}{ViT-B/32} & baseline & 63.82 & 43.44 & 84.63 & 60.22 & 62.44 & 92.17 & 78.05 & 63.71 & 18.78 & 42.17 & 60.94 \\
    & \cellcolor{gray!20}Ours & \cellcolor{gray!20}\textbf{70.85} & \cellcolor{gray!20}\textbf{50.24} & \cellcolor{gray!20}\textbf{89.64} & \cellcolor{gray!20}\textbf{60.32} & \cellcolor{gray!20}\textbf{66.27} & \cellcolor{gray!20}\textbf{92.74} & \cellcolor{gray!20}\textbf{78.07} & \cellcolor{gray!20}\textbf{65.02} & \cellcolor{gray!20}\textbf{21.84} & \cellcolor{gray!20}\textbf{52.42} & \cellcolor{gray!20}\textbf{64.74} \\
    \midrule
    \multirow{2}{*}{ViT-B/16} & baseline & 67.28 & 44.44 & 87.98 & 65.24 & 65.08 & 92.98 & 83.80 & 62.55 & 23.70 & 41.41 & 63.45 \\
    & \cellcolor{gray!20}Ours & \cellcolor{gray!20}\textbf{81.36} & \cellcolor{gray!20}\textbf{53.96} & \cellcolor{gray!20}\textbf{91.58} & \cellcolor{gray!20}\textbf{66.45} & \cellcolor{gray!20}\textbf{70.39} & \cellcolor{gray!20}\textbf{93.59} & \cellcolor{gray!20}\textbf{84.02} & \cellcolor{gray!20}\textbf{67.77} & \cellcolor{gray!20}\textbf{29.73} & \cellcolor{gray!20}\textbf{61.51} & \cellcolor{gray!20}\textbf{70.04} \\
    \midrule
    \multirow{2}{*}{ViT-L/14} & baseline & 75.88 & 54.85 & 93.02 & 77.71 & 75.84 & 95.62 & 89.20 & 70.13 & 31.86 & 51.64 & 71.58 \\
    & \cellcolor{gray!20}Ours & \cellcolor{gray!20}\textbf{82.01} & \cellcolor{gray!20}\textbf{62.47} & \cellcolor{gray!20}\textbf{94.17} & \cellcolor{gray!20}\textbf{77.75} & \cellcolor{gray!20}\textbf{78.38} & \cellcolor{gray!20}\textbf{95.98} & \cellcolor{gray!20}\textbf{89.26} & \cellcolor{gray!20}\textbf{71.51} & \cellcolor{gray!20}\textbf{35.19} & \cellcolor{gray!20}\textbf{66.54} & \cellcolor{gray!20}\textbf{75.33} \\
    \bottomrule
    \end{tabular}
    \caption{Comparison of performance across different architectures. We report the results of CLIP model with standard prompt templates and our method on different architectures.}
    \label{tab:arch}
\end{table*}


\section{Further Details for Hyperparameters}
\label{sec:hyperparameters}

\begin{figure}[t]
    \centering
    \includegraphics[width=0.85\linewidth]{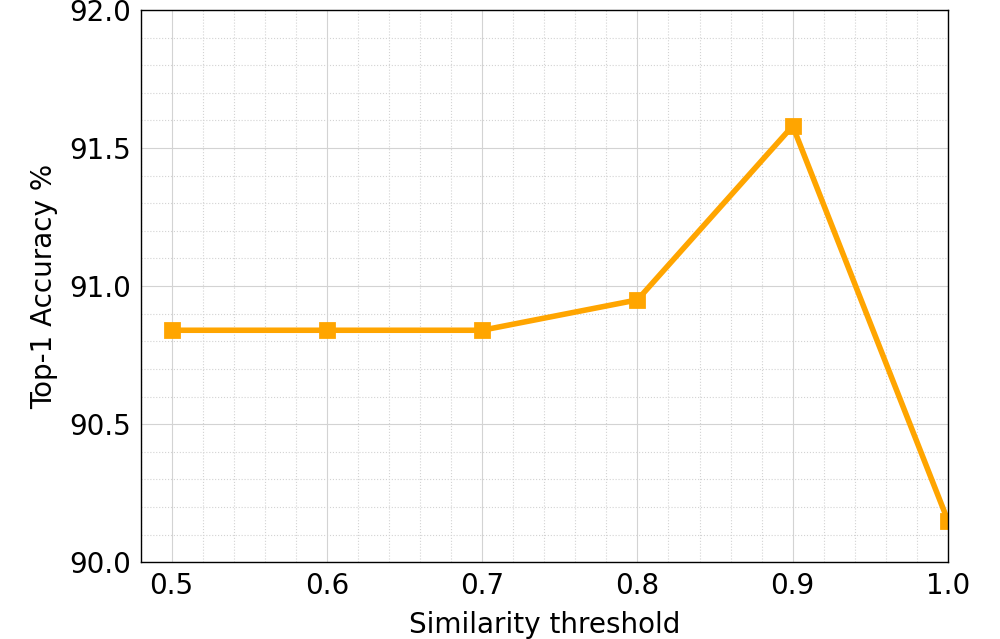}
    \caption{The top-1 accuracy for different similarity thresholds on Pets dataset.}
    \label{fig:threshold}
\end{figure}

\noindent\textbf{Similarity Threshold for Vietoris-Rips Complex.} 
Figure \ref{fig:threshold} illustrates the top-1 accuracy across various similarity thresholds on the Pets dataset. The results demonstrate that increasing the Vietoris-Rips complex similarity threshold leads to significant accuracy improvements, peaking at the threshold of 0.9. However, when the threshold reaches 1.0, the performance drops sharply. This trend suggests that the optimal similarity threshold lies at 0.9. Consequently, we recommend setting the similarity threshold hyperparameter to 0.9 for optimal performance.

\begin{figure}[t]
    \centering
    \includegraphics[width=0.85\linewidth]{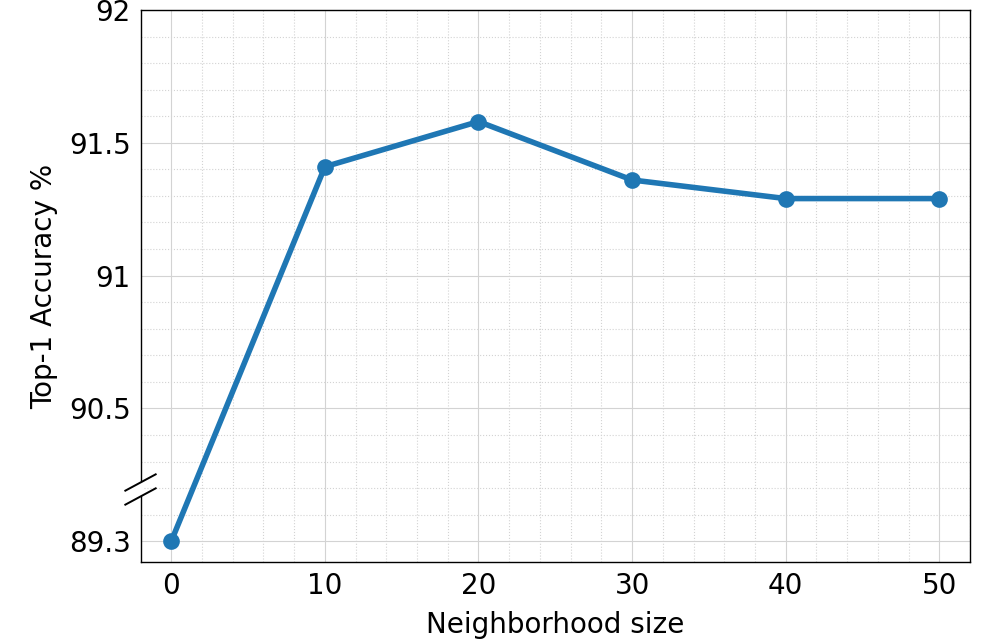}
    \caption{The top-1 accuracy with varying neighborhood sizes on Pets dataset.}
    \label{fig:neighborhood}
\end{figure}

\noindent\textbf{Neighborhood Size for Point-to-Local-Center Metric.} 
Figure \ref{fig:neighborhood} illustrates the top-1 accuracy of the point-to-local-centroid metric across varying neighborhood sizes. The results demonstrate that the accuracy increases rapidly as the neighborhood size grows from 0 to 10, reaching a peak near a size of 20. Beyond this point, the accuracy gradually decreases and stabilizes. Optimal performance is achieved when the neighborhood size ranges from 10 to 30. Therefore, we recommend setting the neighborhood size hyperparameter within this range to achieve the best results.


\section{Detailed Results on Ablation Study}

\noindent\textbf{Selection of LLMs.} 
Table \ref{tab:llm} presents the detailed results of two leading text generation LLMs, GPT-4 and Claude for synonymous texts generation across 10 datasets. While GPT-4 shows a slight advantage over Claude on the Aircraft and EuroSAT datasets, with a margin of 0.3\% and 0.02\%, respectively, Claude outperforms GPT-4 overall, leading by an average of 1.16\% across all 10 datasets.

\noindent\textbf{Effect of Homology.} 
Table \ref{tab:homology} presents the detailed results of the synonymous semantic spaces with \textit{w/} and \textit{w/o} homology across 10 datasets. The addition of homology consistently improves performance across all datasets, with an average increase of 1.54\%. Notably, the improvements are particularly significant on the Flowers, Pets and EuroSAT datasets, with increases of 2.23\%, 2.18\% and 8.72\%, respectively.


\section{Generalizing Across Architectures} 
We evaluate the performance of our method with baseline (CLIP model with standard prompt templates) across varying architectures, specifically ViT-B/32, ViT-B/16, and ViT-L/14. As shown in Table \ref{tab:arch}, our method consistently outperforms baselines across all architectures. Notably, with the ViT-L/14 model, our method achieves the highest average accuracy of 75.33\%, reinforcing its robustness across different architectures.

\end{document}